\documentclass[conference, a4paper]{IEEEtran}
\usepackage{times}
\usepackage{epsfig}
\usepackage{graphicx}
\usepackage{amsmath}
\usepackage{amssymb}
\usepackage{graphicx}
\usepackage{amsmath}
\usepackage{amssymb}
\usepackage{booktabs}
\usepackage{enumerate}
\usepackage{colortbl}
\usepackage{float}
\usepackage{algpseudocode}
\usepackage[ruled]{algorithm2e}
\usepackage{caption}
\usepackage{subcaption}
\usepackage{multirow}
\usepackage{enumitem}
\usepackage{etoolbox}
\usepackage[colorlinks]{hyperref}
\usepackage{cleveref}
\newcommand*{\eg}{e.g.\@\xspace}
\newcommand*{\ie}{i.e.\@\xspace}

\DeclareMathOperator*{\argmax}{argmax}

\ifCLASSINFOpdf
\else
\fi
\begin{document}
%
\title{\textit{Do You Really Mean That?} Content Driven Audio-Visual Deepfake Dataset and Multimodal Method for Temporal Forgery Localization}


\author{\IEEEauthorblockN{Zhixi~Cai}
\IEEEauthorblockA{Monash University\\
zhixi.cai@monash.edu}
\and
\IEEEauthorblockN{Kalin~Stefanov}
\IEEEauthorblockA{Monash University\\
kalin.stefanov@monash.edu}
\and
\IEEEauthorblockN{Abhinav Dhall}
\IEEEauthorblockA{Indian Institute of Technology Ropar\\
Monash University\\
abhinav@iitrpr.ac.in}
\and
\IEEEauthorblockN{Munawar Hayat}
\IEEEauthorblockA{Monash University\\
munawar.hayat@monash.edu}
}



%
\author{\IEEEauthorblockN{Zhixi~Cai\IEEEauthorrefmark{1},
Kalin~Stefanov\IEEEauthorrefmark{1},
Abhinav~Dhall\IEEEauthorrefmark{2}\IEEEauthorrefmark{1} and
Munawar~Hayat\IEEEauthorrefmark{1}
}
\IEEEauthorblockA{
\{zhixi.cai,kalin.stefanov,munawar.hayat\}@monash.edu, abhinav@iitrpr.ac.in}
\IEEEauthorblockA{\IEEEauthorrefmark{1}Monash University, Australia}
\IEEEauthorblockA{\IEEEauthorrefmark{2}Indian Institute of Technology Ropar, India
}
}

\makeatletter
\patchcmd{\@maketitle}
  {\addvspace{-0.3\baselineskip}\egroup}
  {}
  {}
\makeatother

\maketitle
\newcommand\dataset{Localized Audio Visual DeepFake}
\newcommand\datasetabbr{LAV-DF}
\newcommand\model{Boundary Aware Temporal Forgery Detection}
\newcommand\modelabbr{BA-TFD}
\newcommand\task{Temporal Forgery Localization}
\newcommand\tasklower{temporal forgery localization}
\newcommand\taskabbr{TFL}
\begin{abstract}
Due to its high societal impact, deepfake detection is getting active attention in the computer vision community. Most deepfake detection methods rely on identity, facial attributes, and adversarial perturbation-based spatio-temporal modifications at the whole video or random locations while keeping the meaning of the content intact. However, a sophisticated deepfake may contain only a small segment of video/audio manipulation, through which the meaning of the content can be, for example, completely inverted from a sentiment perspective. We introduce a content-driven audio-visual deepfake dataset, termed \dataset{} (\datasetabbr{}), explicitly designed for the task of learning temporal forgery localization. Specifically, the content-driven audio-visual manipulations are performed strategically to change the sentiment polarity of the whole video. Our baseline method for benchmarking the proposed dataset is a 3DCNN model, termed as \model{} (\modelabbr{}), which is guided via contrastive, boundary matching, and frame classification loss functions. Our extensive quantitative and qualitative analysis demonstrates the proposed method's strong performance for temporal forgery localization and deepfake detection tasks.
\end{abstract}

\section{Introduction}
\label{sec:introduction}
Advances in computer vision and deep learning methods (\eg Autoencoders~\cite{rumelhart_learning_1985} and Generative Adversarial Networks~\cite{goodfellow_generative_2020}) have enabled the creation of very realistic fake videos, known as \textit{deepfakes}\footnote{In the text, \textit{deepfake} and \textit{forgery} are used interchangeably.}. There are various ways of creating deepfakes, including voice cloning~\cite{wang_tacotron_2017, jia_transfer_2018}, face reenactment~\cite{tulyakov_mocogan_2018, prajwal_lip_2020}, and face swapping~\cite{korshunova_fast_2017, nirkin_fsgan_2019}. Highly realistic deepfakes are a potential tool for spreading harmful misinformation, given our increasing online presence. This success in generating high-quality deepfakes has raised serious concerns about their role in shaping people's beliefs, with some scholars suggesting that deepfakes are a ``threat to democracy''~\cite{schwartz_you_2018, brandon_there_2019, sample_what_2020, thomas_deepfakes_2020}. As an example of the potentially harmful effect of deepfakes, consider the recent work~\cite{thies_neural_2020} that uses a video of the former United States President Barack Obama to showcase a novel face reenactment method. In this work, the lip movements of Barack Obama are synchronized with another person's speech, resulting in high quality and realistic video in which the former president appears to say something he never did. Given the recent surge in synthesized fake video content on the Internet, it has become increasingly important to identify deepfakes with more accurate and reliable methods. This has led to the release of several benchmark datasets~\cite{korshunov_deepfakes_2018, rossler_faceforensics_2019, dolhansky_deepfake_2020} and methods~\cite{mirsky_creation_2021} for fake content detection. These fake video detection methods aim to correctly classify any given input video as either \textit{real} or \textit{fake}. This suggests that the major assumption behind those datasets and methods is that fake content is present in the entirety of the video/audio signal; that is, there is some form of manipulation throughout the content. And current state-of-the-art deepfake detection methods~\cite{coccomini_combining_2022, heo_deepfake_2021, wodajo_deepfake_2021} achieve impressive results on this problem using the largest benchmark datasets.

\begin{figure}[t]
\centering
\includegraphics[width=\columnwidth]{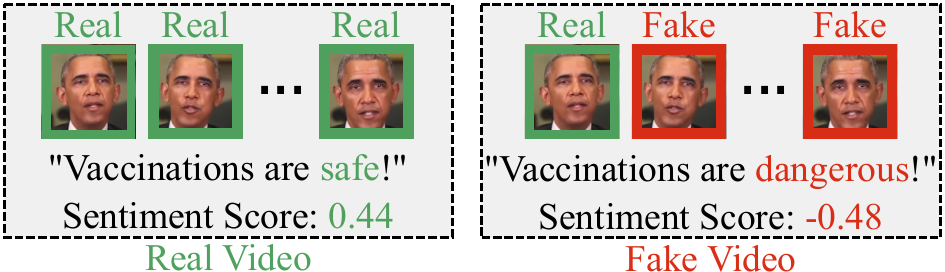}
\caption{\textbf{Content-driven audio-visual manipulation.} On the left is a real video with the subject saying ``Vaccinations are safe''. On the right is an audio-visual deepfake created from the real video based on the change in perceived sentiment where ``safe'' is changed to ``dangerous''. Green-edge and red-edge images are real and fake frames, respectively. \textit{Through subtle audio-visual manipulation, the whole meaning of the video content has changed.}}
\label{fig:overview} 
\vspace{-5mm}
\end{figure}

\begin{table*}[t]
\centering
\caption{\textbf{Comparison of the proposed dataset with other publicly available deepfake datasets.} \textit{Cla: Classification}, \textit{SL: Spatial Localization}, \textit{\taskabbr{}: \task{}}, \textit{FS: Face Swapping}, and \textit{RE: ReEnactment}.}
\label{tab:datasets}
\scalebox{0.85}{
\begin{tabular}{|l|c|c|c|c|c|c|c|c|c|}
\hline
\textbf{Dataset} & \textbf{Year} & \textbf{Tasks} & \textbf{Manipulated} & \textbf{Manipulation} & \textbf{\#Subjects} & \textbf{\#Real} & \textbf{\#Fake} & \textbf{\#Total} \\
&  &  & \textbf{Modality} & \textbf{Method} &  &  &  & \\
\hline\hline
DF-TIMIT~\cite{korshunov_deepfakes_2018} & 2018 & Cla & V &  FS & 43 & 320 & 640 & 960 \\
UADFV~\cite{yang_exposing_2019} & 2019 & Cla & V & FS & 49 & 49 & 49 & 98  \\
FaceForensics++~\cite{rossler_faceforensics_2019} & 2019 & Cla & V & FS/RE & - & 1,000 & 4,000 & 5,000 \\
Google DFD~\cite{nick_contributing_2019} & 2019 & Cla & V & FS & - & 363 & 3,068 & 3,431 \\
DFDC~\cite{dolhansky_deepfake_2020} & 2020 & Cla & AV & FS & 960 & 23,654 & 104,500 & 128,154 \\
DeeperForensics~\cite{jiang_deeperforensics-10_2020} & 2020 & Cla & V & FS & 100 & 50,000 & 10,000 & 60,000 \\
Celeb-DF~\cite{li_celeb-df_2020} & 2020 & Cla & V & FS & 59 & 590 & 5,639 & 6,229 \\
WildDeepfake~\cite{zi_wilddeepfake_2020} & 2021 & Cla & - & - & - & 3,805 & 3,509 & 7,314 \\
FakeAVCeleb~\cite{khalid_fakeavceleb_2021} & 2021 & Cla & AV & RE & 600$+$ & 570 & 25,000$+$ & 25,500$+$ \\
ForgeryNet~\cite{he_forgerynet_2021} & 2021 & SL/\taskabbr{}/Cla & V & Random FS/RE & 5400$+$ & 99,630 & 121,617 & 221,247 \\
\hline
\datasetabbr~(Ours) & 2022 & \taskabbr{}/Cla & AV & Content-driven RE & 153 & 36,431 & 99,873 & 136,304 \\
\hline
\end{tabular}}
\vspace{-5mm}
\end{table*}

However, fake content might constitute only a small part of an otherwise long real video, as was initially suggested in~\cite{chugh_not_2020}. Such short modified segments have the power to alter the meaning and sentiment of the original content completely. For example, consider the manipulation illustrated in Figure~\ref{fig:overview}. The real video might represent a person saying ``Vaccinations are safe'', while the fake includes only a short modified segment; for example, ``safe'' is replaced with ``dangerous''. Hence, the meaning and sentiment of the fake video differ significantly from the real one. If done precisely, this type of coordinated manipulation can sway public opinion (\eg when employed for media of a famous person as the example with Barack Obama) in a particular direction, for example, based on target sentiment polarity. Given the discussed central assumption behind current datasets and methods, the state-of-the-art deepfake detectors might not perform well on this type of manipulation.

This paper tackles the important task of detecting content altering fake segments in videos. The literature review on benchmark datasets for deepfake detection indicates that there is no dataset suitable for this task, that is, a dataset that consists of content-driven manipulations. Therefore, this paper describes the process of creating such a large-scale dataset that will enable further research in this important direction. In addition, we propose a novel multimodal method for precisely predicting the boundaries of fake segments based on visual and audio information. The \textbf{main contributions} of our work are as follows,

\begin{enumerate}
\item{We introduce a new large-scale public audio-visual dataset called \textit{\dataset{}}.}
\item{We propose a new multimodal method called \textit{\model{}}.}
\end{enumerate}

\section{Related Work}
\label{sec:related_work}
\noindent \textbf{Deepfake Datasets.} The body of research in deepfake detection is driven by seminal datasets curated with different manipulation methods. A summary of the relevant datasets is presented in Table~\ref{tab:datasets}. Korshunov and Marcel~\cite{korshunov_deepfakes_2018} curated one of the first deepfake datasets, DF-TIMIT, where face-swapping was performed on VidTimit~\cite{sanderson_vidtimit_2009}. Down the lane, other important datasets such as UADFV~\cite{yang_exploring_2018}, FaceForensics++~\cite{rossler_faceforensics_2019}, and Google DFD~\cite{nick_contributing_2019} were introduced. Due to the complexity of face manipulation and limited availability of open-source face manipulation techniques, these datasets are fairly small in size~\cite{li_celeb-df_2020}. Facebook released a large-scale dataset DFDC~\cite{dolhansky_deepfake_2020} in 2020 for the task of deepfake classification. Multiple face manipulation methods generated 128,154 videos, including real videos of 3000 actors. DFDC has become a mainstream benchmark dataset for the task of deepfake detection. With the progress in both audio and visual deepfake manipulation, post DFDC, several new datasets including Celeb-DF~\cite{li_celeb-df_2020}, DeeperForensics~\cite{jiang_deeperforensics-10_2020}, and WildDeepFake~\cite{zi_wilddeepfake_2020} were introduced. All these datasets are designed for the binary task of deepfake classification and focus primarily on visual manipulation detection~\cite{chugh_not_2020}. In 2021, OpenForensics~\cite{le_openforensics_2021} dataset was introduced for spatial detection, segmentation and classification. Recently, FakeAVCeleb~\cite{khalid_fakeavceleb_2021} was released, focusing on both face-swap and face-reenactment methods with manipulated audio and video. ForgeryNet\cite{he_forgerynet_2021} is the latest contribution to the growing list of deepfake detection datasets. This large-scale dataset is also centered around video-only identity manipulation and is suitable for video/image classification and spatial/temporal forgery localization tasks.

\begin{figure*}[t]
\centering
\vspace{-5mm}
\includegraphics[width=\textwidth]{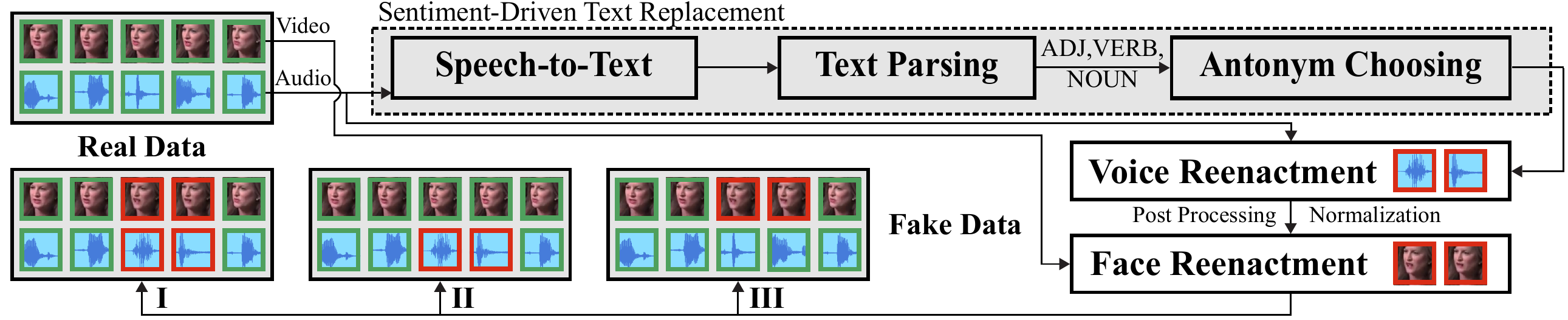}
\caption{\textbf{Generation pipeline of the proposed dataset}. The green-edge audio and video frames are the real data, and the red-edge audio and video frames are the generated fake data. The real audio-based transcript is used to decide the location and content to be replaced based on the largest change in sentiment. The chosen antonyms are used as input for generating fake audio with voice cloning. The post-processing and normalization are applied to the audio to maintain the consistency of the loudness between the generated audio and real audio in the neighborhood. The generated audio is used as input for facial reenactment. Three categories of data are generated: \textit{I. Fake Audio and Fake Video, II. Fake Audio and Real Video} and \textit{III. Real Audio and Fake Video}. The details on dataset generation are discussed in Section~\ref{sec:proposed_dataset}.}
\label{fig:data_generation}
\vspace{-5mm}
\end{figure*}

All previous datasets provide face manipulations that occur in most of the frames of the video~\cite{chugh_not_2020}. Only the latest one, ForgeryNet, provides examples of the important problem of \tasklower{} since it includes random face-swapping applied to parts of some videos. However, the manipulations present in that dataset are only identity modifications that do not necessarily alter the meaning of the content. Our content-driven manipulation dataset addresses this important gap.

\noindent \textbf{Deepfake Detection.} Deepfake detection methods draw inspiration from observations of artifacts such as different eye colors and unnatural blink and lip-sync issues in deepfake videos. These binary classification approaches are based on both traditional machine learning methods (\eg EM~\cite{guarnera_deepfake_2020} and SVM~\cite{yang_exposing_2019}) and deep learning methods (\eg 3DCNN\cite{de_lima_deepfake_2020}, GRU\cite{montserrat_deepfakes_2020} and ViT~\cite{wodajo_deepfake_2021, heo_deepfake_2021, coccomini_combining_2022}). Previous methods~\cite{lewis_deepfake_2020, gu_spatiotemporal_2021} also aim to detect temporal inconsistencies in deepfake content and recently, several audio-visual deepfake detection methods such as MDS~\cite{chugh_not_2020} and M2TR~\cite{wang_m2tr_2021} were proposed. The methods above are classification centric and do not focus on temporal localization. The only exception is the MDS, shown to work for localization tasks, however, the method is designed primarily for classification. The proposed dataset and method are specifically designed for temporal localization of manipulations.

\noindent \textbf{Temporal Localization.} Given that the task of temporal forgery localization is similar to the task of temporal action localization, previous work in this area is important. Benchmark datasets in this domain include THUMOS~\cite{idrees_thumos_2017} and ActivityNet~\cite{caba_heilbron_activitynet_2015} and the proposed methods can be grouped into two categories: 2-step approaches which first generate segment proposals and then perform multi-class classification to evaluate the proposals~\cite{zeng_graph_2019, xu_g-tad_2020, liu_multi-shot_2021} and 1-step approaches which directly generate the final segment predictions~\cite{lin_single_2017,buch_end--end_2019,nawhal_activity_2021}. For temporal forgery localization, there are no classification requirements for the foreground segments; the background is always real, and the foreground segments are always fake. Therefore, boundary prediction and 1-step approaches are more relevant for our task. Bagchi et al.~\cite{bagchi_hear_2021} divided the approaches to segment proposal estimation in temporal action localization into two main categories: methods based on anchors and methods based on predicting the boundary probabilities. As for the anchor-based, these methods mainly use sliding windows in the video, such as S-CNN~\cite{shou_temporal_2016}, CDC~\cite{shou_cdc_2017}, TURN-TAP~\cite{gao_turn_2017} and CTAP~\cite{gao_ctap_2018}. As for the methods predicting the boundary probabilities, Lin et al.~\cite{lin_bsn_2018} introduced BSN. The method can utilize the global information to overcome the problem that anchor-based methods cannot generate precise and flexible segment proposals. Based on BSN, BMN~\cite{lin_bmn_2019} and BSN++~\cite{su_bsn_2021} were introduced for improved performance. It is worth noting that all these methods are unimodal, which is not optimal for the task of temporal forgery localization. The importance of multimodality was demonstrated recently by AVFusion~\cite{bagchi_hear_2021}.
    
\noindent \textbf{Proposed Approach.} For the task of temporal forgery localization, both the audio and visual information are important, in addition to the required precise boundary proposals. In this paper, we introduce a multimodal method based on boundary probabilities and compare its performance with BMN~\cite{lin_bmn_2019}, AGT~\cite{nawhal_activity_2021}, MDS~\cite{chugh_not_2020} and AVFusion~\cite{bagchi_hear_2021}.

\section{Proposed Dataset}
\label{sec:proposed_dataset}
The proposed dataset \dataset{} (\datasetabbr{}) is a large audio-visual deepfake dataset. The main steps in creating the dataset are 1) Sourcing the real videos, 2) Processing the real videos to manipulate their transcripts, and 3) Audio and video synthesis. The deepfake generation is based on the hypothesis that changing relevant words in a transcript can lead to a change in its perception, and in particular, this can be accomplished by changing the sentiment of the transcript. Therefore, the manipulation strategy is to replace strategic words with their antonyms, which leads to a significant change in the sentiment of the statement. The data generation pipeline is illustrated in Figure~\ref{fig:data_generation}.

\noindent \textbf{Data Sourcing.} The real videos are sourced from the VoxCeleb2~\cite{chung_voxceleb2_2018} dataset, a facial video dataset with over 1 million utterance videos of over 6000 speakers. The faces in the videos are tracked and cropped with the facial detector in~\cite{king_dlib-ml_2009} at 224$\times$ 224 resolution. The original dataset contains videos with different duration, spoken language, and voice loudness. Only English-speaking videos are chosen using the confidence score from the Google Speech-to-Text service. The same service generates the transcripts, which are used for manipulation.

\subsection{Data Generation}
\begin{figure}[t]
\centering
\includegraphics[width=\columnwidth]{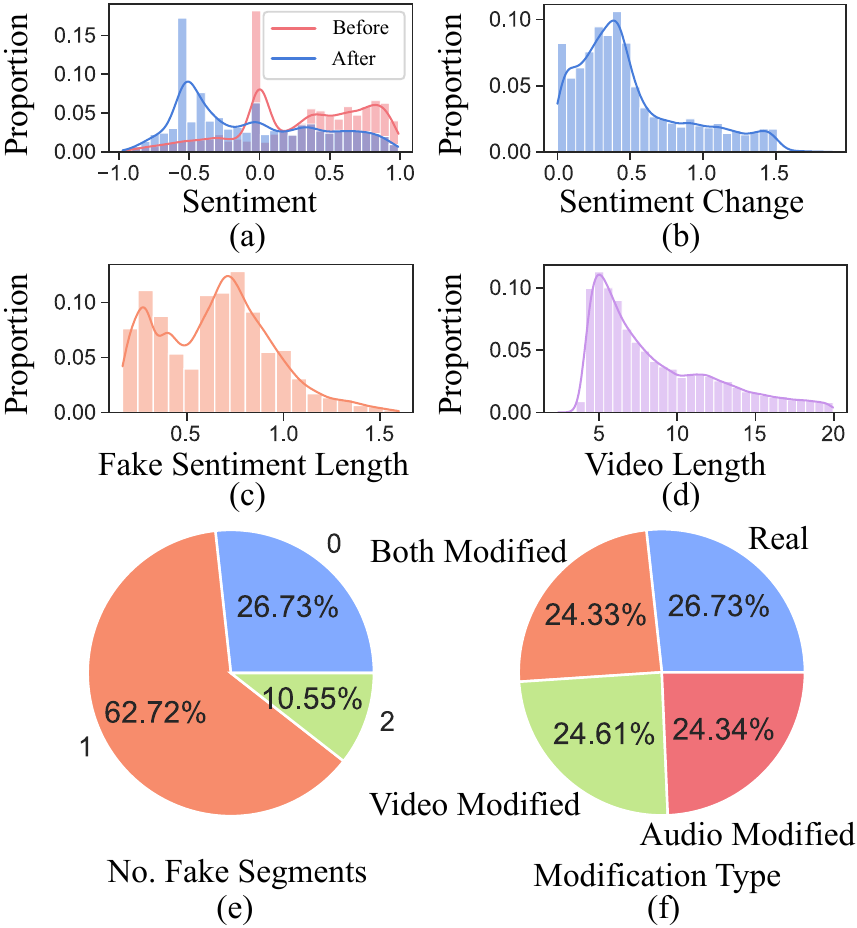}
\caption{\textbf{Summary of the proposed dataset.} (a) Distribution of sentiment scores, (b) Distribution of sentiment changes, (c) Distribution of fake segment lengths, (d) Distribution of video lengths, (e) Proportion of fake segments, and (f) Proportion of modifications.}
\label{fig:data_distribution}
\vspace{-5mm}
\end{figure}

\noindent \textbf{Transcript Manipulation.} After sourcing the real videos, the next step is to analyse a video's transcript denoted by $D = \{d_0, d_1, \cdots, \boldsymbol{d_m}, \cdots, d_n\} $, where $d_i$ denotes word tokens and $n$ is the number of tokens. The aim is to find the tokens to be replaced in $D$ such that the sentiment of the transcript changes the most. In other words, to goal is to create a transcript $D' = \{d_0, d_1, \cdots, \boldsymbol{d_m'}, \cdots, d_n\}$, composed of most of the tokens of $D$ with the exception of a few tokens being replaced. The replacement token $d'_m$ is selected from a set $\hat{d_m}$ of antonyms of $d_m$. The sentiment analyzer in NLTK~\cite{bird_natural_2009} is used to estimate the sentiment value  $S(D)$ of a transcript. For each token $d_i$ in a transcript $D$, we find the replacement as follows,
$$\tau = \argmax_{d_i\in D, d'_i \in \hat{d_i}}|S(D)-S(D')|$$
We find all the replacements in a transcript $D$ as follows,
$$\theta = \argmax_{\{\tau_{m}\}_{m=1}^{M}}|\sum_{i=1}^{M}\Delta S(\tau_{i})|$$
where $\Delta S(\tau_{i})$ is the sentiment difference with the replacement $\tau_{i}$ and $M$ is the maximum number of replacements in the transcript. For videos shorter than 10 seconds, there is up to 1 replacement; otherwise, there are up to 2 replacements. Figure~\ref{fig:data_distribution}~(a) illustrates the change in sentiment distribution after the manipulations and Figure~\ref{fig:data_distribution}~(b) presents the histogram of $|\Delta S|$, suggesting that the sentiment of most transcripts was successfully changed.

\noindent \textbf{Audio Generation.} The next step is to generate the corresponding audio in the speaker’s style. Several recent adaptive text-to-speech (TTS)  methods~\cite{jia_transfer_2018, casanova_sc-glowtts_2021, neekhara_expressive_2021} which can generate the speech style of a person who is not in the training dataset were evaluated. Based on the better performance, SV2TTS~\cite{jia_transfer_2018} is chosen as the final method for audio generation. The SV2TTS comprises three modules 1) An encoder for extracting style embedding of the reference speaker, 2) Tacotron 2~\cite{shen_natural_2018} based spectrogram generated using the replacement tokens and the speaker style embedding, and 3) WaveNet~\cite{oord_wavenet_2016} based vocoder for generating realistic audio using the spectrogram. The pre-trained SV2TTS is used for generating the fake audio segments which are later loudness normalized using the corresponding real audio neighbors.

\begin{figure*}[t]
\centering
\vspace{-5mm}
\includegraphics[width=\textwidth]{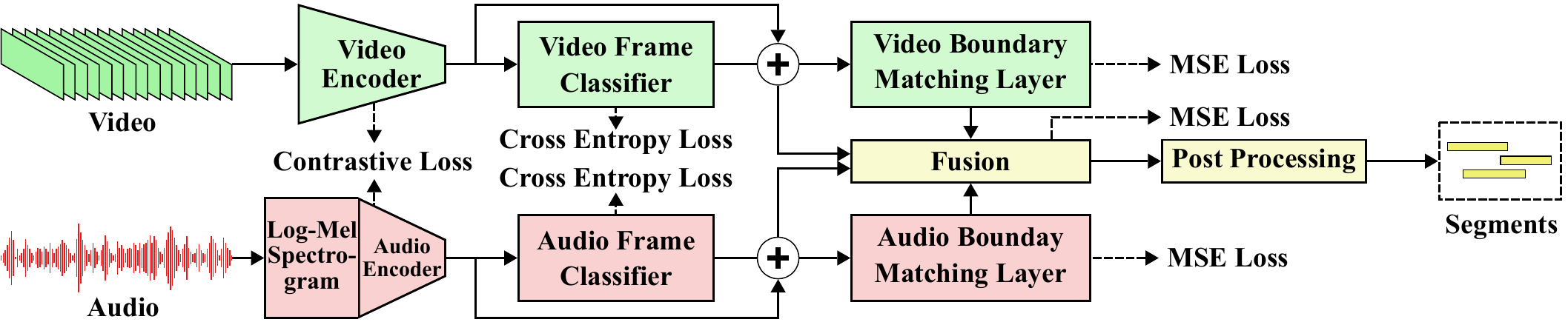}
\caption{\textbf{Structure of the proposed method.} The video encoder uses raw video as input. The audio encoder uses spectrograms extracted from raw audio. $\oplus$ denotes concatenation. During inference, post-processing is applied to generate segments from the output of the fusion module. The details on different components of the method are discussed in Section~\ref{sec:proposed_method}.}
\label{fig:model}
\vspace{-5mm}
\end{figure*}

\noindent \textbf{Video Generation.} The generated fake audio is used as an input for generating the corresponding fake video frames. Wav2Lip~\cite{prajwal_lip_2020} facial reenactment is used for this task as it has been shown to have better output quality than previous methods~\cite{jamaludin_you_2019, k_r_towards_2019}, and has better generalization and robustness to unseen scenarios. It is worth noting that newer methods that achieve better video synthesis quality are not suitable for our task. For example, AD-NeRF~\cite{guo_ad-nerf_2021} is not designed for zero-shot generation of unseen identities, and ATVGNet~\cite{chen_hierarchical_2019} reenacts the face based on a static reference image, which causes pose inconsistencies on the boundary between fake and real segments. Wav2Lip takes a reference video and target audio as input, generating an output video in which the person in the reference video speaks the target audio content with synced lips. The pre-trained Wav2Lip model is used and the generated fake video segments are up-scaled to a resolution of $224\times 224$. In the final step, the generated fake audio and video segments are synchronized and used to replace the original audio and video segments.

Similar to~\cite{khalid_evaluation_2021} the proposed dataset includes three variations of deepfake data,

\begin{enumerate}
\item{\textbf{Fake Audio} and \textbf{Fake Video.} The audio and corresponding video are generated for replacement tokens.}
\item{\textbf{Fake Audio} and \textbf{Real Video.} Only the audio is generated for replacement tokens and the corresponding real video is length-normalized.}
\item{\textbf{Real Audio} and \textbf{Fake Video.} Only the video is generated for replacement tokens and the length of the fake video is normalized to match the real audio.}
\end{enumerate}

\subsection{Dataset Statistics}
The dataset contains 136,304 videos, of which 36,431 are completely real, and 99,873 have fake segments, with 153 unique identities. We split the dataset into 3 identity-independent subsets for training (78703 videos of 91 identities), validation (31501 videos of 31 identities), and testing (26100 videos of 31 identities). The summary of the dataset is shown in Figure~\ref{fig:data_distribution}. The total number of fake segments is 114,253, with duration in the range [0-1.6] seconds and an average length of 0.65 seconds, where 89.26\% of the segments are shorter than 1 second. The maximum video length is 20 seconds, and 69.61\% of the videos are shorter than 10 seconds. As for the modality modification types, the amount of the 4 types (\ie video-modified, audio-modified, both-modified, real) is approximately equal. In most videos (62.72\%), there is 1 fake segment, and in some videos (10.55\%), there are 2. 

\section{Proposed Method}
\label{sec:proposed_method}
The proposed method called \model{} (\modelabbr{}) is illustrated in Figure~\ref{fig:model}. The first step of the method is to extract features from the input data $X = \{V, A\}$, where $V$ is the video and $A$ is the audio.

\subsection{Feature Encoders}
\noindent \textbf{Video Encoder.} The goal of the video encoder is to learn frame-level spatio-temporal features from the input video $V$ using a 3DCNN. For that purpose, we designed the video encoder $E_v$ to take the whole video $V \in \mathbb{R}^{C\times T\times H\times W}$ as input, where $T$ is the number of frames, $C$ is the number of channels, and $H$ and $W$ are the height and width of the frame. The output of the $E_v$ are the frame-level features $F_v \in \mathbb{R}^{C_f \times T}$, where $C_f$ is the features dimension. $E_v$ is composed of 4 blocks, each containing multiple 3D convolutional layers with kernel size $3\times 3\times 3$ and a final max-pooling layer.

\noindent \textbf{Audio Encoder.} The goal of the audio encoder is to learn features from the input audio $A$ using a 2DCNN. In addition, the learned audio features are temporarily aligned with the learned video frame-level video features. The first step is to generate the spectrogram $A^\prime \in \mathbb{R}^{F_m\times T_a}$ of the audio signal in log-space, where $T_a$ is the temporal dimension, and $F_m$ is the length of mel-frequency cepstrum features. In the second step, we designed the audio encoder $E_a$ to take the spectrogram $A^\prime$ as input. The output of the $E_a$ are the audio frame features $F_a \in \mathbb{R}^{C_f \times T}$, where $C_f$ is the features dimension. $E_a$ is composed of multiple 2D convolutional layers with kernel size $3\times 3$ and a final max-pooling layer to reduce the temporal dimension $T_a$ to $T$.

\subsection{Loss Functions}  
\noindent \textbf{Contrastive Loss.} We hypothesize that content modification in one or more modalities will result in miss-synchronization between the modalities (\ie video and audio), and contrastive loss has been shown~\cite{chung_out_2017, chugh_not_2020} to be a powerful objective for similar tasks. Our method uses the audio and video features learned from real videos as positive pairs. The audio and video features learned from videos with at least one modified modality are considered negative pairs. For the positive pairs, the contrastive loss minimizes the difference between the modalities, while for negative pairs, the contrastive loss keeps that margin larger than $\delta$,
$$L_c = \frac{1}{NC_fT}\sum^N_{i=1}y_{i}d_{i}^{2} + (1-y_i)\max(\delta - d_{i}, 0)^2 $$
$$d_i = ||F_{vi} - F_{ai}||_2$$

\noindent \textbf{Frame Classification Loss.} Since we have access to the frame-level features $F_v$ and $F_a$, we utilize the labels and train the encoders to extract powerful and robust features that capture different deepfake artifacts. For that purpose, we designed two frame-level logistic regression classifiers $FC_v$ and $FC_a$ using $F_v$ and $F_a$ as input. The classifiers consist of 1D convolutional layers and predict the label $\hat{Y}$ as real or fake for each frame and each modality. The classifiers are trained with cross-entropy loss,
$$ L_{f} = -\frac{1}{2NT}\sum_{m\in\{a, v\}}\sum^N_{i=1}\sum^T_{j=t}H(\hat{Y}_{mij},Y_{mij}) $$
$$ H(\hat{Y}, Y) = Y\log{\hat{Y}} + (1 - Y)\log{(1-\hat{Y})} $$
$$Y_m = \eta_m Y + (1 - \eta_m)Y_0$$
where $N$ is the number of samples in the dataset, $T$ is the number of frames, $m$ is the modality (\ie audio $a$ or video $v$), $\eta_m$ specifies whether modality $m$ is modified, and $Y_0$ is the label for real videos.

\noindent \textbf{Boundary Matching Loss.} The ground truth boundary maps are generated following the procedure in~\cite{lin_bmn_2019}. Given the fusion boundary map $\hat{M}$, video boundary map $\hat{M_v}$ and audio boundary map $\hat{M_a}$ predicted by the model we use mean squared error as boundary matching loss for $\hat{M}$, $\hat{M_v}$ and ${\hat{M_a}}$. The fusion boundary matching loss is,
$$L_b = \frac{1}{NDT}\sum^N_{i=1}\sum^D_{j=1}\sum^T_{k=1}(\hat{M}_{ijk}-M_{ijk})^2$$
where, $N$ is the number of samples in the dataset, $D$ is the number of all possible proposal durations and $T$ is the number of frames. The modality boundary matching loss is similar to the frame classification loss,
$$L_{bm} = \frac{1}{2NDT}\sum_{m\in\{v,a\}}\sum^N_{i=1}\sum^D_{j=1}\sum^T_{k=1}(\hat{M}_{mijk} - M_{mijk})^2 $$
$$M_m = \eta_m M + (1 - \eta_m)M_0$$
where, $m$ is the modality (\ie video $v$ or audio $a$), $\eta_m$ specifies whether modality $m$ is modified, and $M_0$ is the ground truth boundary map for real videos.

\noindent \textbf{Overall Loss.} The overall loss is defined as follows,
$$L = \lambda_c L_c + \lambda_f L_f + \lambda_b L_b + \lambda_{bm} L_{bm}$$
where, $\lambda_c$, $\lambda_f$, $\lambda_b$ and $\lambda_{bm}$ are weights for different losses.

\begin{figure}[t]
\centering
\includegraphics[width=\columnwidth]{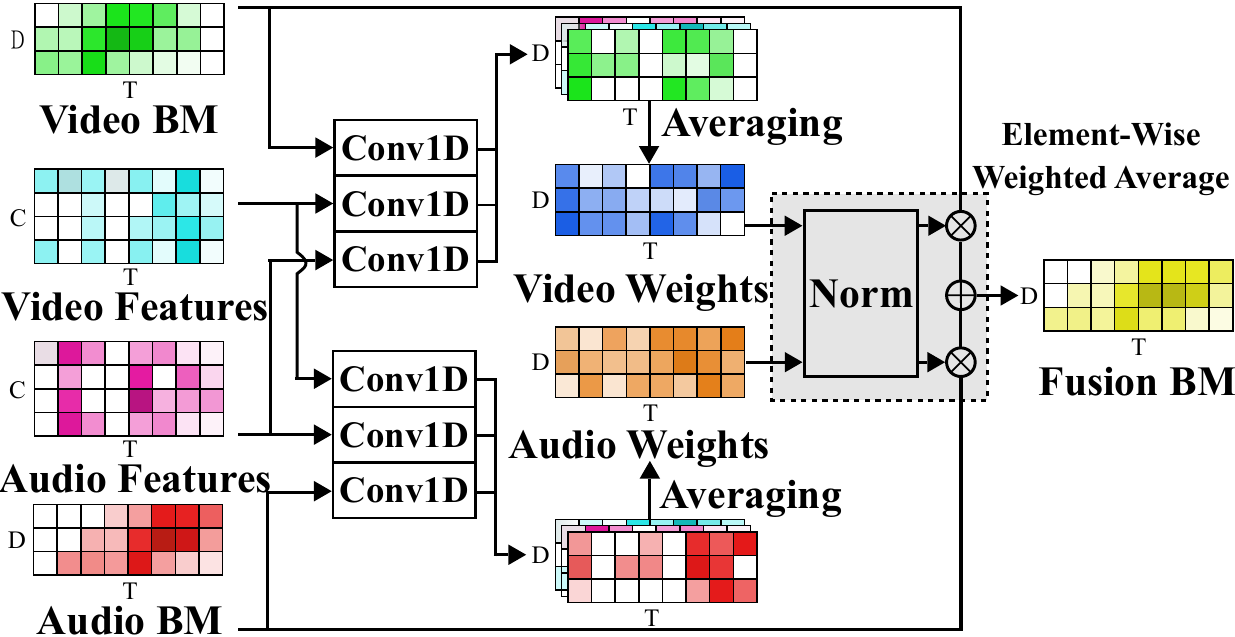}
\caption{\textbf{Structure of the fusion module.} The gray block normalizes the video and audio weights predicted from the 1D convolutional layers and applies element-wise weighted average. $\oplus$ denotes element-wise addition and $\otimes$ denotes element-wise multiplication. \textit{BM: boundary map}.}
\label{fig:fusion}
\vspace{-5mm}
\end{figure}

\subsection{Multimodal Fusion}
The predictions of $FC_v$ and $FC_a$ are concatenated with the features $F_v$ and $F_a$, and used by two boundary matching layers $B_v$ and $B_a$~\cite{lin_bmn_2019}. The goal is to predict the boundary maps $\hat{M}_v \in \mathbb{R}^{D\times T}$ and $\hat{M}_a \in \mathbb{R}^{D\times T}$ for the video and audio, where $T$ is the number of frames and $D$ is the maximum duration of the fake segments. The fusion module, illustrated in Figure~\ref{fig:fusion}, uses the $\hat{M}_v$, $\hat{M}_a$, $F_v$ and $F_a$ as input. For the video modality, the $\hat{M}_v$, $F_v$ and $F_a$ are used to calculate the video weights $W_v \in \mathbb{R}^{D\times T}$ and for the audio modality, the $\hat{M}_a$, $F_a$ and $F_v$ are used to calculate the audio weights $W_a \in \mathbb{R}^{D\times T}$. In the final step, we perform element-wise weighted average and calculate the fusion boundary map prediction $\hat{M} \in \mathbb{R}^{D\times T}$,
$$\hat{M} = \frac{W_v \hat{M}_v + W_a \hat{M}_a}{W_v + W_a}$$
where all operations are element-wise.

\begin{table*}
\centering
\caption{\textbf{Temporal forgery localization results on the full set (see Section~\ref{sec:experiments} for details) of the proposed dataset.} The visual-only version of the proposed method uses the output from the video boundary matching layer (see Figure~\ref{fig:model} for details), showing the performance when using only the video modality.}
\label{tab:fullset}
\scalebox{0.85}{
\begin{tabular}{|l|c|c|c|c|c|c|c|}
\hline
\textbf{Method} & \textbf{AP@0.5} & \textbf{AP@0.75} & \textbf{AP@0.95} & \textbf{AR@100} & \textbf{AR@50} & \textbf{AR@20} & \textbf{AR@10} \\ \hline\hline
MDS~\cite{chugh_not_2020} & 12.78 & 01.62 & 00.00 & 37.88 & 36.71 & 34.39 & 32.15 \\
AGT~\cite{nawhal_activity_2021} & 17.85 & 09.42 & 00.11 & 43.15 & 34.23 & 24.59 & 16.71 \\
BMN~\cite{lin_bmn_2019} & 24.01 & 07.61 & 00.07 & 53.26 & 41.24 & 31.60 & 26.93 \\
BMN (I3D) & 10.56 & 01.66 & 00.00 & 48.49 & 44.39 & 37.13 & 31.55 \\
AVFusion~\cite{bagchi_hear_2021} & 65.38 & 23.89 & 00.11 & 62.98 & 59.26 & 54.80 & 52.11 \\
\hline
\modelabbr{}~(visual-only)~(Ours) & 58.55 & 28.60 & 00.16 & 62.49 & 58.77 & 53.86 & 50.29 \\
\modelabbr{}~(multimodal)~(Ours) & \textbf{76.90} & \textbf{38.50} & \textbf{00.25} & \textbf{66.90} & \textbf{64.08} & \textbf{60.77} & \textbf{58.42} \\
\hline
\end{tabular}}
\end{table*}

\begin{table*}
\centering
\caption{\textbf{Temporal forgery localization results on the subset (see Section~\ref{sec:experiments} for details) of the proposed dataset.} The visual-only version of the proposed method uses the output from the video boundary matching layer (see Figure~\ref{fig:model} for details), showing the performance when using only the video modality.}
\label{tab:subset}
\scalebox{0.85}{
\begin{tabular}{|l|c|c|c|c|c|c|c|}
\hline
\textbf{Method} & \textbf{AP@0.5} & \textbf{AP@0.75} & \textbf{AP@0.95} & \textbf{AR@100} & \textbf{AR@50} & \textbf{AR@20} & \textbf{AR@10} \\ \hline\hline
MDS~\cite{chugh_not_2020} & 23.43 & 03.48 & 00.00 & 58.53 & 56.68 & 53.16 & 49.67 \\
AGT~\cite{nawhal_activity_2021} & 15.69 & 10.69 & 00.15 & 49.11 & 40.31 & 31.70 & 23.13 \\
BMN~\cite{lin_bmn_2019} & 32.32 & 11.38 & 00.14 & 59.69 & 48.17 & 39.01 & 34.17 \\
BMN (I3D) & 28.10 & 05.47 & 00.01 & 55.49 & 54.44 & 52.14 & 47.72 \\
AVFusion~\cite{bagchi_hear_2021} & 62.01 & 22.77 & 00.11 & 61.98 & 58.08 & 53.31 & 50.52 \\
\hline
\modelabbr{}~(visual-only)~(Ours) & 83.55 & 41.88 & 00.24 & 65.79 & 62.30 & 57.95 & 55.34 \\
\modelabbr{}~(multimodal)~(Ours) & \textbf{85.20} & \textbf{47.06} & \textbf{00.29} & \textbf{67.34} & \textbf{64.52} & \textbf{61.19} & \textbf{59.32} \\
\hline
\end{tabular}}
\vspace{-5mm}
\end{table*}    

\subsection{Inference}
During inference, the model uses the video and audio as input and generates a fusion boundary map $\hat{M}$. The boundary map represents the confidence for all proposals in the video and is very dense (\ie there are many duplicated proposals). Similar to BSN~\cite{lin_bsn_2018}, we utilize post-processing with Soft Non-Maximum Suppression (S-NMS)~\cite{bodla_soft-nms_2017} to eliminate the duplicated proposals.

\section{Experiments}
\label{sec:experiments}
We have performed extensive benchmarking of the proposed dataset via several state-of-the-art methods including, BMN~\cite{lin_bmn_2019}, AGT~\cite{nawhal_activity_2021}, AVFusion~\cite{bagchi_hear_2021}, and MDS~\cite{chugh_not_2020}. Apart from our proposed dataset, we also validate our method for classification on DFDC~\cite{dolhansky_deepfake_2020} dataset.

\noindent \textbf{Dataset Preparation and Evaluation Protocol.} To compare with visual-only methods, we prepare a subset of the test set where the audio-only modified data is removed which is denoted as \textit{subset}. The original test set is denoted as \textit{full set} in the experiments. Unlike temporal action localization methods~\cite{liu_multi-shot_2021, nawhal_activity_2021} that are using only average precision, we follow the protocol proposed in  ForgeryNet~\cite{he_forgerynet_2021} and use both average precision (AP) and average recall (AR) as the evaluation metrics for the quantitative comparison. For AP, we follow the protocol of ActivityNet~\cite{caba_heilbron_activitynet_2015} to set the IoU thresholds to 0.5, 0.75 and 0.95. For AR, as the number of fake segments is small, we set the number of proposals to 100, 50, 20 and 10 with the IoU thresholds [0.5:0.05:0.95]. Our method can also be used for deepfake detection (\ie classification) task. We use area under the curve (AUC) for evaluation of the deepfake classification.

\noindent \textbf{Implementation Details.} The proposed method is implemented in PyTorch~\cite{paszke_pytorch_2019}. For hyperparameters, we set $\lambda_c$ = $0.1$, $\lambda_f$ = $2$, $\lambda_b$ = $1$, $\lambda_{bm}$ = $1$ and $\delta$ = $0.99$. For comparison, we trained BMN~\cite{lin_bmn_2019}, AGT~\cite{nawhal_activity_2021}, AVFusion~\cite{bagchi_hear_2021} and MDS~\cite{chugh_not_2020} for temporal forgery localization task. In addition, to evaluate the usefulness of the proposed method, we compare with MDS, EfficientViT~\cite{coccomini_combining_2022} and other methods on classification task. We followed the original settings for BMN, AGT, MDS and EfficientViT, and used encoding concatenation fusion for AVFusion. For the methods that require pre-trained features, we trained them end-to-end with trainable encoder. For comparison, we also trained BMN with I3D features~\cite{carreira_quo_2017} (\ie fixed encoder). For the models which require S-NMS~\cite{bodla_soft-nms_2017} post-processing, we used the validation set to search for optimal parameters for post-processing. Final evaluation and results are based on the test set. For DFDC, we consider the whole fake video as one fake segment. For evaluation, we used 2 methods to generate the classification output for our method 1) Using the highest confidence of the predicted segments as the confidence of the video being fake and 2) Training a MLP classifier using the confidences of predicted segments. We chose evaluation method 1) for our dataset and method 2) for DFDC based on performance.

\begin{table*}
\centering
\caption{\textbf{Temporal forgery localization results on the full set (see Section~\ref{sec:experiments} for details) of the proposed dataset.} The contribution of different loss terms in the proposed method (see Section~\ref{sec:proposed_method} for details).}
\label{tab:losses}
\scalebox{0.85}{
\begin{tabular}{|l|c|c|c|c|c|c|c|}
\hline
\textbf{Loss Function} & \textbf{AP@0.5} & \textbf{AP@0.75} & \textbf{AP@0.95} & \textbf{AR@100} & \textbf{AR@50} & \textbf{AR@20} & \textbf{AR@10} \\ 
\hline\hline
$L_f$ & 40.50 & 29.74 & 00.13 & 60.51 & 60.50 & 60.47 & 59.90 \\
$L_c, L_f$ & 40.92 & 31.23 & 00.74 & 64.71 & 64.71 & 64.36 & 62.79 \\
$L_b$ & 53.16 & 11.91 & 00.02 & 53.99 & 50.94 & 47.74 & 45.55 \\
$L_{bm}, L_b$ & 54.70 & 15.50 & 00.04 & 56.64 & 53.57 & 49.46 & 45.85 \\
$L_{f}, L_{bm}, L_b$ & 76.50 & \textbf{39.92} & 00.18 & 66.69 & 63.71 & 60.07 & 57.76 \\
$L_{c}, L_{f}, L_{bm}, L_b$ & \textbf{76.90} & 38.50 & \textbf{00.25} & \textbf{66.90} & \textbf{64.08} & \textbf{60.77} & \textbf{58.42} \\
\hline
\end{tabular}}
\vspace{-5mm}
\end{table*}
    
\section{Results}
\label{sec:results}
\noindent \textbf{Temporal Forgery Localization.} We compare our method on the full set of the proposed dataset with the latest methods for temporal action localization and deepfake detection. From Table~\ref{tab:fullset}, our method achieves the best performance, which is 76.9 for AP@0.5 and 66.9 for AR@100. Unlike temporal action localization datasets, in our dataset there is a single label for the fake segments, so it is reasonable that the AP score is relatively high. The multimodal MDS method is not designed for temporal forgery localization tasks and predicts only fixed length segments (\ie cannot predict the precise boundaries), hence the scores for that method are low. As for AGT and BMN, the scores are low because they are visual-only unimodal methods and cannot detect the fake segments in videos where only the audio is modified. We also evaluated the performance of our visual-only unimodal method, which shows worse results than the multimodal version and AVFusion. In addition, the results show that when the video encoder is trained with data from the proposed dataset, BMN performs significantly better than using I3D features. We also evaluated the same methods on the subset of the proposed dataset. From Table~\ref{tab:subset}, the performance of the visual-only methods is improved, and for our method, the visual-only score improves from 58.55~(AP@0.5) to 83.55~(AP@0.5) and the margin between the unimodal and multimodal versions is decreased from 18.35~(AP@0.5) to 1.65~(AP@0.5). Overall, our method still ranks first, which demonstrates it's superior performance for temporal forgery detection.

\noindent \textbf{Deepfake Classification.} We also compare our method with previous deepfake detection methods on the full set of the proposed dataset and a subset of DFDC. On our dataset, our method (\textbf{0.990}) outperforms F3Net~\cite{qian_thinking_2020} (0.520), MDS (0.828) and EfficientViT (0.965). As for the subset of DFDC, the performance of our method (\textbf{0.846}) is better than previous methods such as Meso4~\cite{afchar_mesonet_2018} (0.753), FWA~\cite{li_exposing_2019} (0.727) and \cite{mittal_emotions_2020} (0.844) and is close to MDS (0.916). It is worth noting that, our method is not designed and trained for classification task with classification loss. It is trained for temporal forgery localization and then the segment outputs are summarized as a whole video label prediction. Therefore, the performance of our method on DFDC drops as compared to the state-of-the-art classification method MDS. On the other hand, previous deepfake detection methods assume that fake videos are entirely fake, so their performance (\eg the frame-based approach of F3Net) is reduced on our dataset. In summary, our method still performs well on classification task and has potential to reach the state-of-the-art performance.

\noindent \textbf{Impact of Loss Functions.} From Table~\ref{tab:losses}, all loss terms have positive effect on the performance of the proposed model. The results suggest that the frame classification loss contributes the most to the method performance. 

\noindent \textbf{Failure Analysis.} The output of the proposed method can be noisy for cases that contain very short video manipulations ($\leq 0.5$ sec) and the corresponding real audio. For such short video-only manipulations, if the visual transition from real to fake and then back to real is smooth, it may lead to noisy output.

\section{Conclusion} This work introduces and investigates a novel problem related to content-driven deepfake generation and detection. To this end, we propose a new dataset in which the audio and video are modified at specific locations based on the change in sentiment of the content. We also propose a new method for temporal forgery localization in such partially modified videos. The conducted experiments show that our method achieves better performance than previous relevant state-of-the-art methods.

\noindent \textbf{Ethical Concerns.} 
The proposed dataset potentially might have a negative social impact. Since the individuals in the dataset are celebrities, the content in the dataset may be used for unethical purposes such as making fake rumours. Also, the dataset generation pipeline can be used to create fake videos. To encounter the potential negative impact of our work, we prepared a license for public usage of the dataset and proposed the method.

\noindent \textbf{Limitations.} This work has some limitations 1) The audio reenactment method used in the dataset does not always generate the reference style, 2) The resolution of the dataset is constrained on the basis of source videos and 3) The high score of classification results indicates the necessity of improving the video reenactment method. 

\noindent \textbf{Future Work.} Major improvement in the future will be increasing the dataset with new token insertion, substitution and deletion of existing tokens and converting statements into questions.

\bibliographystyle{IEEEtran}
\bibliography{paper}

\begin{thebibliography}{10}
\providecommand{\url}[1]{#1}
\csname url@samestyle\endcsname
\providecommand{\newblock}{\relax}
\providecommand{\bibinfo}[2]{#2}
\providecommand{\BIBentrySTDinterwordspacing}{\spaceskip=0pt\relax}
\providecommand{\BIBentryALTinterwordstretchfactor}{4}
\providecommand{\BIBentryALTinterwordspacing}{\spaceskip=\fontdimen2\font plus
\BIBentryALTinterwordstretchfactor\fontdimen3\font minus
  \fontdimen4\font\relax}
\providecommand{\BIBforeignlanguage}[2]{{%
\expandafter\ifx\csname l@#1\endcsname\relax
\typeout{** WARNING: IEEEtran.bst: No hyphenation pattern has been}%
\typeout{** loaded for the language `#1'. Using the pattern for}%
\typeout{** the default language instead.}%
\else
\language=\csname l@#1\endcsname
\fi
#2}}
\providecommand{\BIBdecl}{\relax}
\BIBdecl

\bibitem{rumelhart_learning_1985}
D.~E. Rumelhart, G.~E. Hinton, and R.~J. Williams,
  ``\BIBforeignlanguage{en}{Learning {Internal} {Representations} by {Error}
  {Propagation}},'' Tech. Rep., 1985.

\bibitem{goodfellow_generative_2020}
I.~Goodfellow, J.~Pouget-Abadie, M.~Mirza, B.~Xu, D.~Warde-Farley, S.~Ozair,
  A.~Courville, and Y.~Bengio, ``Generative adversarial networks,''
  \emph{Communications of the ACM}, vol.~63, no.~11, pp. 139--144, 2020.

\bibitem{wang_tacotron_2017}
Y.~Wang, R.~J. Skerry-Ryan, D.~Stanton, and et~al., ``Tacotron: {Towards}
  {End}-to-{End} {Speech} {Synthesis},'' \emph{arXiv:1703.10135 [cs]}, 2017.

\bibitem{jia_transfer_2018}
Y.~Jia, Y.~Zhang, R.~J. Weiss, and et~al., ``Transfer learning from speaker
  verification to multispeaker text-to-speech synthesis,'' in
  \emph{NeurIPS}.\hskip 1em plus 0.5em minus 0.4em\relax Curran Associates
  Inc., 2018, pp. 4485--4495.

\bibitem{tulyakov_mocogan_2018}
S.~Tulyakov, M.-Y. Liu, X.~Yang, and J.~Kautz, ``{MoCoGAN}: {Decomposing}
  {Motion} and {Content} for {Video} {Generation},'' in \emph{CVPR}, 2018, pp.
  1526--1535.

\bibitem{prajwal_lip_2020}
K.~R. Prajwal, R.~Mukhopadhyay, V.~P. Namboodiri, and C.~Jawahar, ``A {Lip}
  {Sync} {Expert} {Is} {All} {You} {Need} for {Speech} to {Lip} {Generation}
  {In} the {Wild},'' in \emph{ACM MM}, 2020, pp. 484--492.

\bibitem{korshunova_fast_2017}
I.~Korshunova, W.~Shi, J.~Dambre, and L.~Theis, ``Fast {Face}-{Swap} {Using}
  {Convolutional} {Neural} {Networks},'' in \emph{ICCV}, 2017, pp. 3677--3685.

\bibitem{nirkin_fsgan_2019}
Y.~Nirkin, Y.~Keller, and T.~Hassner, ``{FSGAN}: {Subject} {Agnostic} {Face}
  {Swapping} and {Reenactment},'' in \emph{ICCV}, 2019, pp. 7184--7193.

\bibitem{schwartz_you_2018}
O.~Schwartz, ``\BIBforeignlanguage{en-GB}{You thought fake news was bad? {Deep}
  fakes are where truth goes to die},'' \emph{\BIBforeignlanguage{en-GB}{The
  Guardian}}, 2018.

\bibitem{brandon_there_2019}
J.~Brandon, ``\BIBforeignlanguage{en}{There {Are} {Now} 15,000 {Deepfake}
  {Videos} on {Social} {Media}. {Yes}, {You} {Should} {Worry}.}''
  \emph{\BIBforeignlanguage{en}{Forbes}}, 2019.

\bibitem{sample_what_2020}
I.~Sample, ``\BIBforeignlanguage{en-GB}{What are deepfakes – and how can you
  spot them?}'' \emph{\BIBforeignlanguage{en-GB}{The Guardian}}, 2020.

\bibitem{thomas_deepfakes_2020}
D.~Thomas, ``\BIBforeignlanguage{auto}{Deepfakes: {A} threat to democracy or
  just a bit of fun?}'' \emph{\BIBforeignlanguage{auto}{BBC News}}, 2020.

\bibitem{thies_neural_2020}
J.~Thies, M.~Elgharib, A.~Tewari, C.~Theobalt, and M.~Nießner,
  ``\BIBforeignlanguage{en}{Neural {Voice} {Puppetry}: {Audio}-{Driven}
  {Facial} {Reenactment}},'' in \emph{\BIBforeignlanguage{en}{{ECCV} 2020}},
  2020, pp. 716--731.

\bibitem{korshunov_deepfakes_2018}
P.~Korshunov and S.~Marcel, ``{DeepFakes}: a {New} {Threat} to {Face}
  {Recognition}? {Assessment} and {Detection},'' \emph{arXiv:1812.08685 [cs]},
  2018.

\bibitem{rossler_faceforensics_2019}
A.~Rossler, D.~Cozzolino, L.~Verdoliva, C.~Riess, J.~Thies, and M.~Niessner,
  ``{FaceForensics}++: {Learning} to {Detect} {Manipulated} {Facial}
  {Images},'' in \emph{ICCV}, 2019, pp. 1--11.

\bibitem{dolhansky_deepfake_2020}
B.~Dolhansky, J.~Bitton, B.~Pflaum, J.~Lu, R.~Howes, M.~Wang, and C.~C. Ferrer,
  ``The {DeepFake} {Detection} {Challenge} ({DFDC}) {Dataset},''
  \emph{arXiv:2006.07397 [cs]}, 2020.

\bibitem{mirsky_creation_2021}
Y.~Mirsky and W.~Lee, ``The {Creation} and {Detection} of {Deepfakes}: {A}
  {Survey},'' \emph{ACM Computing Surveys}, vol.~54, no.~1, pp. 7:1--7:41,
  2021.

\bibitem{coccomini_combining_2022}
D.~A. Coccomini, N.~Messina, C.~Gennaro, and F.~Falchi,
  ``\BIBforeignlanguage{en}{Combining {EfficientNet} and {Vision}
  {Transformers} for {Video} {Deepfake} {Detection}},'' in
  \emph{\BIBforeignlanguage{en}{{ICIAP} 2022}}, 2022, pp. 219--229.

\bibitem{heo_deepfake_2021}
Y.-J. Heo, Y.-J. Choi, Y.-W. Lee, and B.-G. Kim, ``Deepfake {Detection}
  {Scheme} {Based} on {Vision} {Transformer} and {Distillation},''
  \emph{arXiv:2104.01353 [cs]}, 2021.

\bibitem{wodajo_deepfake_2021}
D.~Wodajo and S.~Atnafu, ``Deepfake {Video} {Detection} {Using} {Convolutional}
  {Vision} {Transformer},'' \emph{arXiv:2102.11126 [cs]}, 2021.

\bibitem{yang_exposing_2019}
X.~Yang, Y.~Li, and S.~Lyu, ``Exposing {Deep} {Fakes} {Using} {Inconsistent}
  {Head} {Poses},'' in \emph{{ICASSP}}, 2019, pp. 8261--8265.

\bibitem{nick_contributing_2019}
D.~Nick and J.~Andrew, ``\BIBforeignlanguage{en}{Contributing {Data} to
  {Deepfake} {Detection} {Research}},'' 2019.

\bibitem{jiang_deeperforensics-10_2020}
L.~Jiang, R.~Li, W.~Wu, C.~Qian, and C.~C. Loy, ``{DeeperForensics}-1.0: {A}
  {Large}-{Scale} {Dataset} for {Real}-{World} {Face} {Forgery} {Detection},''
  in \emph{CVPR}, 2020, pp. 2889--2898.

\bibitem{li_celeb-df_2020}
Y.~Li, X.~Yang, P.~Sun, and et~al., ``Celeb-{DF}: {A} {Large}-{Scale}
  {Challenging} {Dataset} for {DeepFake} {Forensics},'' in \emph{CVPR}, 2020,
  pp. 3207--3216.

\bibitem{zi_wilddeepfake_2020}
B.~Zi, M.~Chang, J.~Chen, and et~al., ``{WildDeepfake}: {A} {Challenging}
  {Real}-{World} {Dataset} for {Deepfake} {Detection},'' in \emph{ACM MM},
  2020, pp. 2382--2390.

\bibitem{khalid_fakeavceleb_2021}
H.~Khalid, S.~Tariq, and S.~S. Woo, ``{FakeAVCeleb}: {A} {Novel}
  {Audio}-{Video} {Multimodal} {Deepfake} {Dataset},'' \emph{arXiv:2108.05080
  [cs]}, 2021.

\bibitem{he_forgerynet_2021}
Y.~He, B.~Gan, S.~Chen, Y.~Zhou, G.~Yin, L.~Song, L.~Sheng, J.~Shao, and
  Z.~Liu, ``\BIBforeignlanguage{en}{{ForgeryNet}: {A} {Versatile} {Benchmark}
  for {Comprehensive} {Forgery} {Analysis}},'' in
  \emph{\BIBforeignlanguage{en}{CVPR}}, 2021, pp. 4360--4369.

\bibitem{chugh_not_2020}
K.~Chugh, P.~Gupta, A.~Dhall, and R.~Subramanian, ``Not made for each other-
  {Audio}-{Visual} {Dissonance}-based {Deepfake} {Detection} and
  {Localization},'' in \emph{ACM MM}, 2020, pp. 439--447.

\bibitem{sanderson_vidtimit_2009}
C.~Sanderson and B.~C. Lovell, ``Multi-region probabilistic histograms for
  robust and scalable identity inference,'' in \emph{Advances in
  Biometrics}.\hskip 1em plus 0.5em minus 0.4em\relax Berlin, Heidelberg:
  Springer Berlin Heidelberg, 2009, pp. 199--208.

\bibitem{yang_exploring_2018}
K.~Yang, P.~Qiao, D.~Li, S.~Lv, and Y.~Dou, ``\BIBforeignlanguage{en}{Exploring
  {Temporal} {Preservation} {Networks} for {Precise} {Temporal} {Action}
  {Localization}},'' \emph{\BIBforeignlanguage{en}{AAAI}}, vol.~32, no.~1,
  2018.

\bibitem{le_openforensics_2021}
T.-N. Le, H.~H. Nguyen, J.~Yamagishi, and I.~Echizen,
  ``\BIBforeignlanguage{en}{{OpenForensics}: {Large}-{Scale} {Challenging}
  {Dataset} for {Multi}-{Face} {Forgery} {Detection} and {Segmentation}
  {In}-the-{Wild}},'' in \emph{\BIBforeignlanguage{en}{ICCV}}, 2021, pp.
  10\,117--10\,127.

\bibitem{guarnera_deepfake_2020}
L.~Guarnera, O.~Giudice, and S.~Battiato, ``{DeepFake} {Detection} by
  {Analyzing} {Convolutional} {Traces},'' in \emph{CVPRW}, 2020, pp. 666--667.

\bibitem{de_lima_deepfake_2020}
O.~de~Lima, S.~Franklin, S.~Basu, B.~Karwoski, and A.~George, ``Deepfake
  {Detection} using {Spatiotemporal} {Convolutional} {Networks},''
  \emph{arXiv:2006.14749 [cs, eess]}, 2020.

\bibitem{montserrat_deepfakes_2020}
D.~M. Montserrat, H.~Hao, and et~al., ``Deepfakes {Detection} {With}
  {Automatic} {Face} {Weighting},'' in \emph{CVPRW}, 2020, pp. 668--669.

\bibitem{lewis_deepfake_2020}
J.~K. Lewis, I.~E. Toubal, H.~Chen, and et~al., ``Deepfake {Video} {Detection}
  {Based} on {Spatial}, {Spectral}, and {Temporal} {Inconsistencies} {Using}
  {Multimodal} {Deep} {Learning},'' in \emph{AIPR}, 2020, pp. 1--9.

\bibitem{gu_spatiotemporal_2021}
Z.~Gu, Y.~Chen, T.~Yao, S.~Ding, J.~Li, F.~Huang, and L.~Ma, ``Spatiotemporal
  {Inconsistency} {Learning} for {DeepFake} {Video} {Detection},'' in \emph{ACM
  MM}, 2021, pp. 3473--3481.

\bibitem{wang_m2tr_2021}
J.~Wang, Z.~Wu, J.~Chen, and Y.-G. Jiang, ``{M2TR}: {Multi}-modal {Multi}-scale
  {Transformers} for {Deepfake} {Detection},'' \emph{arXiv:2104.09770 [cs]},
  2021.

\bibitem{idrees_thumos_2017}
H.~Idrees, A.~R. Zamir, Y.-G. Jiang, and et~al., ``The {THUMOS} {Challenge} on
  {Action} {Recognition} for {Videos} "in the {Wild}",'' \emph{Computer Vision
  and Image Understanding}, vol. 155, pp. 1--23, 2017.

\bibitem{caba_heilbron_activitynet_2015}
F.~Caba~Heilbron, V.~Escorcia, B.~Ghanem, and J.~Carlos~Niebles,
  ``{ActivityNet}: {A} {Large}-{Scale} {Video} {Benchmark} for {Human}
  {Activity} {Understanding},'' in \emph{CVPR}, 2015, pp. 961--970.

\bibitem{zeng_graph_2019}
R.~Zeng, W.~Huang, M.~Tan, and et~al., ``Graph {Convolutional} {Networks} for
  {Temporal} {Action} {Localization},'' in \emph{ICCV}, 2019, pp. 7094--7103.

\bibitem{xu_g-tad_2020}
M.~Xu, C.~Zhao, D.~S. Rojas, and et~al., ``G-{TAD}: {Sub}-{Graph}
  {Localization} for {Temporal} {Action} {Detection},'' in \emph{CVPR}, 2020,
  pp. 10\,156--10\,165.

\bibitem{liu_multi-shot_2021}
X.~Liu, Y.~Hu, S.~Bai, and et~al., ``\BIBforeignlanguage{en}{Multi-{Shot}
  {Temporal} {Event} {Localization}: {A} {Benchmark}},'' in
  \emph{\BIBforeignlanguage{en}{CVPR}}, 2021, pp. 12\,596--12\,606.

\bibitem{lin_single_2017}
T.~Lin, X.~Zhao, and Z.~Shou, ``Single {Shot} {Temporal} {Action}
  {Detection},'' in \emph{ACM {MM}}, 2017, pp. 988--996.

\bibitem{buch_end--end_2019}
S.~Buch, V.~Escorcia, B.~Ghanem, and et~al., ``\BIBforeignlanguage{English
  (US)}{End-to-end, single-stream temporal action detection in untrimmed
  videos},'' \emph{\BIBforeignlanguage{English (US)}{BMVC}}, 2019.

\bibitem{nawhal_activity_2021}
M.~Nawhal and G.~Mori, ``Activity {Graph} {Transformer} for {Temporal} {Action}
  {Localization},'' \emph{arXiv:2101.08540 [cs]}, 2021.

\bibitem{bagchi_hear_2021}
A.~Bagchi, J.~Mahmood, D.~Fernandes, and R.~K. Sarvadevabhatla, ``Hear {Me}
  {Out}: {Fusional} {Approaches} for {Audio} {Augmented} {Temporal} {Action}
  {Localization},'' \emph{arXiv:2106.14118 [cs]}, 2021.

\bibitem{shou_temporal_2016}
Z.~Shou, D.~Wang, and S.-F. Chang, ``Temporal {Action} {Localization} in
  {Untrimmed} {Videos} via {Multi}-{Stage} {CNNs},'' in \emph{CVPR}, 2016, pp.
  1049--1058.

\bibitem{shou_cdc_2017}
Z.~Shou, J.~Chan, A.~Zareian, K.~Miyazawa, and S.-F. Chang, ``{CDC}:
  {Convolutional}-{De}-{Convolutional} {Networks} for {Precise} {Temporal}
  {Action} {Localization} in {Untrimmed} {Videos},'' in \emph{CVPR}, 2017, pp.
  5734--5743.

\bibitem{gao_turn_2017}
J.~Gao, Z.~Yang, K.~Chen, C.~Sun, and R.~Nevatia, ``{TURN} {TAP}: {Temporal}
  {Unit} {Regression} {Network} for {Temporal} {Action} {Proposals},'' in
  \emph{ICCV}, 2017, pp. 3628--3636.

\bibitem{gao_ctap_2018}
J.~Gao, K.~Chen, and R.~Nevatia, ``{CTAP}: {Complementary} {Temporal} {Action}
  {Proposal} {Generation},'' in \emph{{ECCV}}, 2018, pp. 68--83.

\bibitem{lin_bsn_2018}
T.~Lin, X.~Zhao, H.~Su, C.~Wang, and M.~Yang, ``{BSN}: {Boundary} {Sensitive}
  {Network} for {Temporal} {Action} {Proposal} {Generation},'' in \emph{ECCV},
  2018, pp. 3--19.

\bibitem{lin_bmn_2019}
T.~Lin, X.~Liu, X.~Li, E.~Ding, and S.~Wen, ``{BMN}: {Boundary}-{Matching}
  {Network} for {Temporal} {Action} {Proposal} {Generation},'' in \emph{ICCV},
  2019, pp. 3889--3898.

\bibitem{su_bsn_2021}
H.~Su, W.~Gan, W.~Wu, Y.~Qiao, and J.~Yan, ``{BSN}++: {Complementary}
  {Boundary} {Regressor} with {Scale}-{Balanced} {Relation} {Modeling} for
  {Temporal} {Action} {Proposal} {Generation},'' \emph{arXiv:2009.07641 [cs]},
  2021.

\bibitem{chung_voxceleb2_2018}
J.~S. Chung, A.~Nagrani, and A.~Zisserman, ``{VoxCeleb2}: {Deep} {Speaker}
  {Recognition},'' in \emph{{INTERSPEECH}}, 2018.

\bibitem{king_dlib-ml_2009}
D.~E. King, ``Dlib-ml: {A} {Machine} {Learning} {Toolkit},'' \emph{The Journal
  of Machine Learning Research}, vol.~10, pp. 1755--1758, 2009.

\bibitem{bird_natural_2009}
S.~Bird, E.~Klein, and E.~Loper, \emph{\BIBforeignlanguage{en}{Natural
  {Language} {Processing} with {Python}: {Analyzing} {Text} with the {Natural}
  {Language} {Toolkit}}}.\hskip 1em plus 0.5em minus 0.4em\relax O'Reilly
  Media, Inc., 2009.

\bibitem{casanova_sc-glowtts_2021}
E.~Casanova, C.~Shulby, E.~Gölge, and et~al., ``{SC}-{GlowTTS}: an {Efficient}
  {Zero}-{Shot} {Multi}-{Speaker} {Text}-{To}-{Speech} {Model},''
  \emph{arXiv:2104.05557 [cs, eess]}, 2021.

\bibitem{neekhara_expressive_2021}
P.~Neekhara, S.~Hussain, S.~Dubnov, F.~Koushanfar, and J.~McAuley,
  ``\BIBforeignlanguage{en}{Expressive {Neural} {Voice} {Cloning}},'' in
  \emph{\BIBforeignlanguage{en}{{ACML}}}, 2021, pp. 252--267.

\bibitem{shen_natural_2018}
J.~Shen, R.~Pang, R.~J. Weiss, and et~al., ``Natural {TTS} {Synthesis} by
  {Conditioning} {Wavenet} on {MEL} {Spectrogram} {Predictions},'' in
  \emph{ICASSP}, 2018, pp. 4779--4783.

\bibitem{oord_wavenet_2016}
A.~v.~d. Oord, S.~Dieleman, H.~Zen, and et~al., ``{WaveNet}: {A} {Generative}
  {Model} for {Raw} {Audio},'' \emph{arXiv:1609.03499 [cs]}, 2016.

\bibitem{jamaludin_you_2019}
A.~Jamaludin, J.~S. Chung, and A.~Zisserman, ``\BIBforeignlanguage{en}{You
  {Said} {That}?: {Synthesising} {Talking} {Faces} from {Audio}},''
  \emph{\BIBforeignlanguage{en}{International Journal of Computer Vision}},
  vol. 127, no.~11, pp. 1767--1779, 2019.

\bibitem{k_r_towards_2019}
P.~K~R, R.~Mukhopadhyay, J.~Philip, A.~Jha, V.~Namboodiri, and C.~V. Jawahar,
  ``Towards {Automatic} {Face}-to-{Face} {Translation},'' in \emph{ACM MM},
  2019, pp. 1428--1436.

\bibitem{guo_ad-nerf_2021}
Y.~Guo, K.~Chen, S.~Liang, Y.-J. Liu, H.~Bao, and J.~Zhang,
  ``\BIBforeignlanguage{en}{{AD}-{NeRF}: {Audio} {Driven} {Neural} {Radiance}
  {Fields} for {Talking} {Head} {Synthesis}},'' in
  \emph{\BIBforeignlanguage{en}{ICCV}}, 2021, pp. 5784--5794.

\bibitem{chen_hierarchical_2019}
L.~Chen, R.~K. Maddox, Z.~Duan, and C.~Xu, ``Hierarchical {Cross}-{Modal}
  {Talking} {Face} {Generation} {With} {Dynamic} {Pixel}-{Wise} {Loss},'' in
  \emph{CVPR}, 2019, pp. 7832--7841.

\bibitem{khalid_evaluation_2021}
H.~Khalid, M.~Kim, S.~Tariq, and S.~S. Woo, ``Evaluation of an {Audio}-{Video}
  {Multimodal} {Deepfake} {Dataset} using {Unimodal} and {Multimodal}
  {Detectors},'' \emph{{ADGD}}, pp. 7--15, 2021.

\bibitem{chung_out_2017}
J.~S. Chung and A.~Zisserman, ``\BIBforeignlanguage{en}{Out of {Time}:
  {Automated} {Lip} {Sync} in the {Wild}},'' in
  \emph{\BIBforeignlanguage{en}{{ACCV} {Workshops}}}, 2017, pp. 251--263.

\bibitem{bodla_soft-nms_2017}
N.~Bodla, B.~Singh, R.~Chellappa, and L.~S. Davis, ``Soft-{NMS} -- {Improving}
  {Object} {Detection} {With} {One} {Line} of {Code},'' in \emph{ICCV}, 2017,
  pp. 5561--5569.

\bibitem{paszke_pytorch_2019}
A.~Paszke, S.~Gross, F.~Massa, and et~al., ``{PyTorch}: {An} {Imperative}
  {Style}, {High}-{Performance} {Deep} {Learning} {Library},'' in
  \emph{Advances in {NeurIPS}}, vol.~32, 2019.

\bibitem{carreira_quo_2017}
J.~Carreira and A.~Zisserman, ``Quo {Vadis}, {Action} {Recognition}? {A} {New}
  {Model} and the {Kinetics} {Dataset},'' in \emph{CVPR}, 2017, pp. 6299--6308.

\bibitem{qian_thinking_2020}
Y.~Qian, G.~Yin, L.~Sheng, Z.~Chen, and J.~Shao,
  ``\BIBforeignlanguage{en}{Thinking in {Frequency}: {Face} {Forgery}
  {Detection} by {Mining} {Frequency}-{Aware} {Clues}},'' in
  \emph{\BIBforeignlanguage{en}{{ECCV}}}, 2020, pp. 86--103.

\bibitem{afchar_mesonet_2018}
D.~Afchar, V.~Nozick, J.~Yamagishi, and I.~Echizen, ``{MesoNet}: a {Compact}
  {Facial} {Video} {Forgery} {Detection} {Network},'' in \emph{{WIFS}}, 2018,
  pp. 1--7.

\bibitem{li_exposing_2019}
Y.~Li and S.~Lyu, ``\BIBforeignlanguage{en}{Exposing {DeepFake} {Videos} {By}
  {Detecting} {Face} {Warping} {Artifacts}},'' in
  \emph{\BIBforeignlanguage{en}{{CVPRW}}}, 2019, p.~7.

\bibitem{mittal_emotions_2020}
T.~Mittal, U.~Bhattacharya, R.~Chandra, A.~Bera, and D.~Manocha, ``Emotions
  {Don}'t {Lie}: {An} {Audio}-{Visual} {Deepfake} {Detection} {Method} using
  {Affective} {Cues},'' in \emph{ACM MM}, 2020, pp. 2823--2832.

\end{thebibliography}

\end{document}


%
\title{\textit{Do You Really Mean That?} Content Driven Audio-Visual Deepfake Dataset and Multimodal Method for Temporal Forgery Localization\\
\begin{Large}Supplementary Material\end{Large}
}





%
\author{\IEEEauthorblockN{Zhixi~Cai\IEEEauthorrefmark{1},
Kalin~Stefanov\IEEEauthorrefmark{1},
Abhinav~Dhall\IEEEauthorrefmark{2}\IEEEauthorrefmark{1} and
Munawar~Hayat\IEEEauthorrefmark{1}
}
\IEEEauthorblockA{
\{zhixi.cai,kalin.stefanov,munawar.hayat\}@monash.edu, abhinav@iitrpr.ac.in}
\IEEEauthorblockA{\IEEEauthorrefmark{1}Monash University, Australia}
\IEEEauthorblockA{\IEEEauthorrefmark{2}Indian Institute of Technology Ropar, India
}
}




\maketitle
\newcommand\dataset{Localized Audio Visual DeepFake}
\newcommand\datasetabbr{LAV-DF}
\newcommand\model{Boundary Aware Temporal Forgery Detection}
\newcommand\modelabbr{BA-TFD}
\newcommand\task{Temporal Forgery Localization}
\newcommand\tasklower{temporal forgery localization}
\newcommand\taskabbr{TFL}

\section{Proposed Method Analysis}
\label{sec:proposed_method_analysis}
\noindent \textbf{Complexity Analysis.} The proposed model is trained on a single RTX3090 GPU using the proposed large-scale dataset for 120 hours with batch size 4. Table~\ref{tab:complexity} provides comparison of the proposed method and other related methods in terms of the number of parameters. The results demonstrate that the proposed method has the potential to scale up for better performance.

\begin{table}[h!]
\centering
\caption{\textbf{Method comparison in terms of number of parameters.} \textit{\modelabbr{}: the proposed method.}}
\label{tab:complexity}
\scalebox{1}{
\begin{tabular}{|c|cccc|}
\hline
\textbf{Type} & \multicolumn{2}{c|}{\textbf{Classification}} & \multicolumn{2}{c|}{\textbf{Localization}} \\
\hline\hline
\textbf{Method} & \multicolumn{1}{c|}{\textbf{MDS~\cite{chugh_not_2020}}} &  \multicolumn{1}{c|}{\textbf{EffiencientViT~\cite{coccomini_combining_2022}}} & \multicolumn{1}{c|}{\textbf{BMN~\cite{lin_bmn_2019}}} & \textbf{\modelabbr{}} \\
\hline
\textbf{\#Params} & \multicolumn{1}{c|}{122.8M} & \multicolumn{1}{c|}{109.4M} & \multicolumn{1}{c|}{50.3M} & \multicolumn{1}{c|}{5.5M} \\
\hline
\end{tabular}}
\end{table}

\noindent \textbf{Qualitative Analysis.} We selected 1 real video and the 3 corresponding fake videos and visualized the boundary map outputs $\hat{M}_v$, $\hat{M}_a$ and $\hat{M}$ in Figure~\ref{fig:outputs}. The results illustrates that 1) The video output captures the fake segments when the video modality is modified, 2) The audio output captures the fake segments when the audio modality is manipulated, and 3) Regardless of whether the audio or video modality is modified, the fusion output captures the correct information from audio and visual outputs demonstrating the effectiveness of the fusion module.

\begin{figure}[h!]
\centering
\includegraphics[width=\columnwidth]{image/outputs.pdf}
\caption{\textbf{Visualization of boundary map outputs.} The first column illustrates the modality-wise boundary map outputs for a real video. The rest of the columns illustrate the modality-wise boundary map outputs for the corresponding fake videos. \textit{GT: ground truth} and \textit{mod: modified}.} 
\label{fig:outputs}
\end{figure}

\noindent \textbf{Failure Analysis.} The output of the proposed method can be noisy for cases that contain very short video manipulations ($\leq 0.5$ sec) and the corresponding real audio. For such short video-only manipulations, if the visual transition from real to fake and then back to real is smooth, it may lead to noisy output.

\noindent \textbf{Loss Functions Analysis.} Figure~\ref{fig:features} provides a visualization of the distributions of video features $F_v$ for 10 videos in the dataset. The results illustrate that the combination of the proposed loss terms leads to more separable features contributing to improved performance in temporal localization.

\section{Proposed Dataset Analysis}
\label{sec:proposed_dataset_analysis}
\noindent \textbf{Visual Quality Analysis.} Table~\ref{tab:visual_quality} provides a comparison of the proposed method and other related methods from literature in terms of visual quality, demonstrating that our method achieves better visual quality. The scores on VoxCeleb2 dataset for the related deepfake generation methods are taken from \cite{zhou_pose-controllable_2021} since it presents evaluation of all mentioned methods on that dataset. The scores for our method are calculated using the partially-fake videos in our dataset and the original real videos in the VoxCeleb2 dataset. It is worth noting that 1) We cannot guarantee the complete overlap between the data used to calculate the scores from the related methods and our method and 2) The partially-fake videos in our dataset include real frames, hence the relatively high performance in terms of PSNR and SSIM.

\begin{table*}[th!]
\centering
\caption{\textbf{Data generation comparison in terms of visual quality.} \textit{\datasetabbr{}: the proposed data generation pipeline.}}
\label{tab:visual_quality}
\scalebox{1}{
\begin{tabular}{|c|cccccc|}
\hline
\textbf{Dataset} & \multicolumn{6}{c|}{\textbf{VoxCeleb2~\cite{chung_voxceleb2_2018}}} \\
\hline\hline
\textbf{Method} &  \multicolumn{1}{c|}{\textbf{ATVG~\cite{chen_hierarchical_2019}}} & \multicolumn{1}{c|}{\textbf{Wav2Lip~\cite{prajwal_lip_2020}}} & \multicolumn{1}{c|}{\textbf{MakeitTalk~\cite{zhou_makelttalk_2020}}} &
\multicolumn{1}{c|}{\textbf{Rhythmic Head~\cite{chen_talking-head_2020}}} & \multicolumn{1}{c|}{\textbf{PC-AVS~\cite{zhou_pose-controllable_2021}}} & \textbf{\datasetabbr{}} \\
\hline
\textbf{PSNR} & \multicolumn{1}{c|}{29.41} & \multicolumn{1}{c|}{29.54} & \multicolumn{1}{c|}{29.51} & \multicolumn{1}{c|}{29.55} & \multicolumn{1}{c|}{29.68} & \textbf{33.06} \\
\hline
\textbf{SSIM} & \multicolumn{1}{c|}{0.826} & \multicolumn{1}{c|}{0.846} & \multicolumn{1}{c|}{0.817} & \multicolumn{1}{c|}{0.779} & \multicolumn{1}{c|}{0.886} & \textbf{0.898} \\
\hline
\end{tabular}}
\end{table*}

\begin{figure*}[t!]
\centering
\includegraphics[width=\textwidth]{image/feature_distributions3.pdf}
\caption{\textbf{Feature distribution in PCA subspace.} Each point is the features of a video frame.}
\label{fig:features}
\end{figure*}

\begin{figure*}[t!]
\centering
\begin{subfigure}[b]{\textwidth}
\centering
\includegraphics[width=\textwidth]{image/outputs2.pdf}
\caption{}
\vspace{2mm}
\end{subfigure}
\begin{subfigure}[b]{\textwidth}
\centering
\includegraphics[width=\textwidth]{image/outputs2_2.pdf}
\caption{}
\end{subfigure}
\begin{subfigure}[b]{\textwidth}
\centering
\includegraphics[width=\textwidth]{image/outputs2_3.pdf}
\caption{}
\end{subfigure}
\caption{\textbf{Visualization of boundary map outputs for successful cases for videos with 2 fake segments.} The first column illustrates the modality-wise boundary map output for a real video. The rest of the columns illustrate the modality-wise boundary map output of the corresponding fake videos.}
\label{fig:output_2seg}
\end{figure*}

\begin{figure*}[t!]
\centering
\begin{subfigure}[b]{\textwidth}
\centering
\includegraphics[width=\textwidth]{image/outputs_fail.pdf}
\caption{}
\vspace{2mm}
\end{subfigure}
\begin{subfigure}[b]{\textwidth}
\centering
\includegraphics[width=\textwidth]{image/outputs_fail_1.pdf}
\caption{}
\end{subfigure}
\begin{subfigure}[b]{\textwidth}
\centering
\includegraphics[width=\textwidth]{image/outputs_fail_2.pdf}
\caption{}
\end{subfigure}
\caption{\textbf{Visualization of boundary map outputs for failure cases for video-only manipulations with short fake segments.} Each boundary map is the output from the fusion and boundary matching modules.}
\label{fig:output_fail}
\end{figure*}
\begin{figure*}[h!]
\centering
\begin{subfigure}[b]{\textwidth}
\centering
\includegraphics[width=\textwidth]{image/outputs_notgood.pdf}
\caption{}
\vspace{2mm}
\end{subfigure}
\begin{subfigure}[b]{\textwidth}
\centering
\includegraphics[width=\textwidth]{image/outputs_notgood_2.pdf}
\caption{}
\vspace{2mm}
\label{fig:outputs_notgood_2}
\end{subfigure}
\begin{subfigure}[b]{\textwidth}
\centering
\includegraphics[width=\textwidth]{image/outputs_notgood_3.pdf}
\caption{}
\vspace{2mm}
\label{fig:outputs_notgood_3}
\end{subfigure}
\caption{\textbf{Visualization of boundary map outputs for inaccurate cases.} The first column illustrates the modality-wise boundary map output for a real video. The rest of the columns illustrate the modality-wise boundary map output of the corresponding fake videos.}
\label{fig:output_notgood}
\end{figure*}

\section{Additional Qualitative Analysis}
\label{sec:additional_qualitative_analysis}
\noindent \textbf{Successful Cases.} Figure~\ref{fig:output_2seg} illustrates the boundary map output for 3 successful cases each including 2 modified segments. This qualitative result demonstrates that the proposed model and fusion module can extract and fuse strong representations from different modalities suitable for the task of temporal forgery localization.

\noindent \textbf{Failure Cases.} Figure~\ref{fig:output_fail} illustrates 3 failure cases. From the visualizations, the confidence of the video output is lower compared to the successful cases, which indicates that the proposed method is unable to detect the very short video-only modifications because the transition from real to fake and then to real is smooth.

\noindent \textbf{Inaccurate Cases.} Figure~\ref{fig:output_notgood} illustrates 3 inaccurate cases. We can observe that the real data produces noise in boundary map and the confidence of the predictions for the ground-truth segments are low in video-only and audio-only manipulations which causes the fusion output to be inaccurate. From the observations, the frames in these cases are comparably static without significant motions, so the model might wrongly predict boundaries of the fake segments. Although the video modality generates false positive predictions (in the left part of the video output), the fusion layer suppresses this output to achieve better precision.

\bibliographystyle{IEEEtran}
\bibliography{paper}